\documentclass[letterpaper, 10 pt, conference]{ieeeconf}  

\IEEEoverridecommandlockouts                              

\usepackage{lineno}
\usepackage{cite}
\usepackage{algorithm}
\usepackage{algpseudocode}
\usepackage{amsmath,amssymb,amsfonts}
\usepackage{graphicx} 
\usepackage{epstopdf}
\usepackage{textcomp}
\usepackage{xcolor}
\usepackage{subfigure}
\usepackage{booktabs}
\usepackage{multirow}
\usepackage{mathrsfs}
\usepackage{soul}
\usepackage{bbding}
\usepackage{setspace}
\usepackage{tabu}                     
\usepackage{multirow}                 
\usepackage{multicol}                 
\usepackage{multirow}                
\usepackage{float}                    
\usepackage{makecell}                 
\usepackage{booktabs}                 

\soulregister\cite7
\soulregister\ref7 
\usepackage{comment}
\usepackage[colorlinks,bookmarksopen,bookmarksnumbered,citecolor=black, linkcolor=black, urlcolor=black]{hyperref}

\usepackage{color}

\title{\LARGE \bf Bioinspired Sensing of Undulatory Flow Fields Generated \\by Leg Kicks in Swimming}

\author{Jun Wang, Tongsheng Shen, Dexin Zhao* and Feitian Zhang*
\thanks{J. Wang (wjun@stu.pku.edu.cn) and F. Zhang (feitian@pku.edu.cn) are with the Department of Advanced Manufacturing and Robotics, and the State Key Laboratory of Turbulence and Complex Systems, College of Engineering, Peking University, Beijing 100871, China. J. Wang is also with the National Innovation Institute of Defense Technology, Beijing, China.
}%
\thanks{T. Shen (shents\_bj@126.com) and D. Zhao (zhaodx2008@163.com) are with the National Innovation Institute of Defense Technology, Beijing 100071, China.}
\thanks{The Dataset is available at https://github.com/PKUROBOT/Swimmer-Leg-Kick-Sensing.git.}
\thanks{The videos for complete experimental procedure are available at https://youtube.com/playlist?list=PL1P-pATAer1v8c2ifO2QtfGFWGapIquj D\&si=eae5OV3gp0UmQD5I.}
\thanks{* Send all correspondences to D. Zhao and F. Zhang.}
}

\begin{document}
\maketitle
\begin{abstract}
  The artificial lateral line (ALL) is a bioinspired flow sensing system for underwater robots, comprising of distributed flow sensors. The ALL has been successfully applied to detect the undulatory flow fields generated by body undulation and tail-flapping of bioinspired robotic fish. However, its feasibility and performance in sensing the undulatory flow fields produced by human leg kicks during swimming has not been systematically tested and studied. This paper presents a novel sensing framework to investigate the undulatory flow field generated by swimmer's leg kicks, leveraging bioinspired ALL sensing. To evaluate the feasibility of using the ALL system for sensing the undulatory flow fields generated by swimmer leg kicks, this paper designs an experimental platform integrating an ALL system and a lab-fabricated human leg model. To enhance the accuracy of flow sensing, this paper proposes a feature extraction method that dynamically fuses time-domain and time-frequency characteristics. Specifically, time-domain features are extracted using one-dimensional convolutional neural networks and bidirectional long short-term memory networks (1DCNN-BiLSTM), while time-frequency features are extracted using short-term Fourier transform and two-dimensional convolutional neural networks (STFT-2DCNN). These features are then dynamically fused based on attention mechanisms to achieve accurate sensing of the undulatory flow field. Furthermore, extensive experiments are conducted to test various scenarios inspired by human swimming, such as leg kick pattern recognition and kicking leg localization, achieving satisfactory results.
\end{abstract}

\begin{NTP}
    This paper tackles the challenge of sensing the flow fields created by human leg kicks during swimming using a bioinspired artificial lateral line (ALL) system. Unlike the undulating movement of a fish’s body or tail, human leg kicks vary in frequency and phase, and very limited research has been done on using ALL systems to sense these movements. To address this, we propose a new method for detecting complex flow patterns with the ALL system. We integrate a lab-fabricated human leg model with the ALL system to show how it captures the features of leg kicks. This method improves flow sensing accuracy by combining time-domain and time-frequency features using advanced neural networks. Key findings include the successful recognition of leg kick patterns and precise localization of the kicking leg, demonstrating the ALL system’s potential for sensing the undulatory flow fields. This study aims to provide theoretical and technical insights for human-robot interaction and formation using ALL systems, enhancing the capabilities of underwater robots and swimmers in complex environments. Additionally, this technology could help develop assistive robots for swimmers, boosting human abilities underwater and ensuring safety during aquatic activities.
\end{NTP}
\section{Introduction}

In nature, fish possess a vital sensory organ known as the lateral line, which plays a critical role in their behavior, particularly in complex underwater environments\cite{Partridge_1980_JCP,Zhai_JBE_2021,Liu_ABB_2016}. By sensing flow velocity, direction, and pressure gradient of their surrounding flow field, the lateral line enables fish to gather environmental information, facilitating behaviors such as obstacle avoidance, rheotaxis, and predation\cite{Zhai_JBE_2021,Liu_ABB_2016,Shizhe_MTMN_2014,montgomery1997Nature}. Inspired by this biological mechanism, researchers have developed various artificial lateral line (ALL) systems to enhance underwater sensing capabilities in robots\cite{Liu_2018_MST,Ling_2024_TIM,Zheng_2020_TRO,Wang_2024_ISJ,Liu_2024_Small}. Applications using ALL systems\cite{Kruusmaa_2011_ICRA,Xu_2017_JOE,Salumae_2013_rspa,Zhang_2024_Tase} such as obstacle avoidance\cite{Free_2018_BIB,Li_2024_ISJ,Xie_2023_BIB}, station holding\cite{DeVries_2015_BIB,Zheng_2021_TRO}, and localization\cite{Wolf_2020_JTRSI,Jiang_2022_Tmech}, have been successfully implemented.

Current research on ALL systems has primarily focused on sensing flow fields generated by bioinspired robotic fish or fish-like models with body and/or fin undulations. For instance, Xie \textit{et al.} extensively studied the flow field generated by the tail flapping of robotic fish using ALL sensing, achieving relative state estimation and energy harvesting between a pair of robotic fish\cite{Li_2021_NC,Zheng_2017_BIB,Zheng_2020_BIB}. Yen \textit{et al.} used ALL to sense the flow field generated by the tail flapping of robotic fish, enabling follower robots to locate the leader\cite{Yen_2020_BIB}. Furthermore, Free \textit{et al.} achieved fish-like swimming behavior in a Karman vortex street by sensing the motion of a fish-like robot and combining it with closed-loop control\cite{Free_2018_BIB}. Xu \textit{et al.} utilized deep learning-based wake sensing to achieve lateral position tracking and oscillation phase matching for leader-follower hydrofoils\cite{Xu_2022_BIB}. 

Studies have found that human leg kicks during swimming produce significant undulatory effects in the surrounding flow field\cite{Hochstein_2011_HMS,andersen_2018_JSS,pacholak_2014_SB,zamparo_2002_JEB,soton492330}. For example, Steffen \textit{et al.} identified flow structures in continuous dolphin kicks using numerical simulations, noting that each kick generates a circular vortex that falls into the swimmer's wake\cite{pacholak_2014_SB}. Zamparo \textit{et al.} demonstrated through experiments that bending waves may occur during kicking, with or without fins\cite{zamparo_2002_JEB}. 
Research on the surrounding flow field during human swimming indicates that human kicks, similar to fish tail flapping, create undulatory flow fields with shedding vortices\cite{andersen_2018_JSS}. These findings confirm that leg kicks produce prominent vortices, suggesting significant potential for using the ALL system to sense the undulatory flow field therein.

However, there has been limited exploration of using ALL systems to sense the undulatory flow fields generated by human swimming. Unlike the undulation of a fish's body or tail, human leg kicks vary in frequency and phase between the two legs. For example, there are two common leg kick styles in swimming --- the dolphin kick and the flutter kick\cite{andersen_2018_JSS}. In the dolphin kick, the legs move synchronously, while in the flutter kick, the legs alternate and kick with half a cycle phase delay\cite{zamparo_2002_JEB}. Compared to fish tail flaps, human leg kicks typically generate more complex flow fields.  

The primary objective of this paper is to study the sensing of the undulatory flow field generated by human leg kicks using the ALL system, which holds significant potential for human-robot interaction (HRI) between underwater robots and human swimmers\cite{Chavez_2021_RAM,Birk_2022_CRR}.
To facilitate this study, a laboratory experiment is designed, drawing inspiration from prior studies on ALL sensing in robotic fish \cite{Zheng_2017_BIB,Rodwell_2023_bib,Free_2018_BIB} and human flow field analysis \cite{NAKASHIMA_2010_jbse,nakashima_2007_jfst,nakashima_2009_Jbse}. Two tasks are investigated including leg kick pattern recognition and kicking leg localization.

The contribution of this paper is twofold.
First, to the best of the authors' knowledge, this paper investigates, for the first time, the sensing of undulatory flow fields generated by human leg kicks using the ALL system, proposing a bioinspired sensing framework. This study holds significant implications for future research in HRI between underwater robots and human swimmers. 
Second, to accurately sense the undulatory flow field, a fusion algorithm based on the attention mechanism is proposed to assimilate time-domain and time-frequency features. 
Furthermore, various HRI-inspired scenarios, including leg kick pattern recognition and kicking leg localization, are experimentally tested, yielding satisfactory results.

\section{Problem Description}
\begin{figure}
    \centering
    \includegraphics[width=0.951\linewidth]{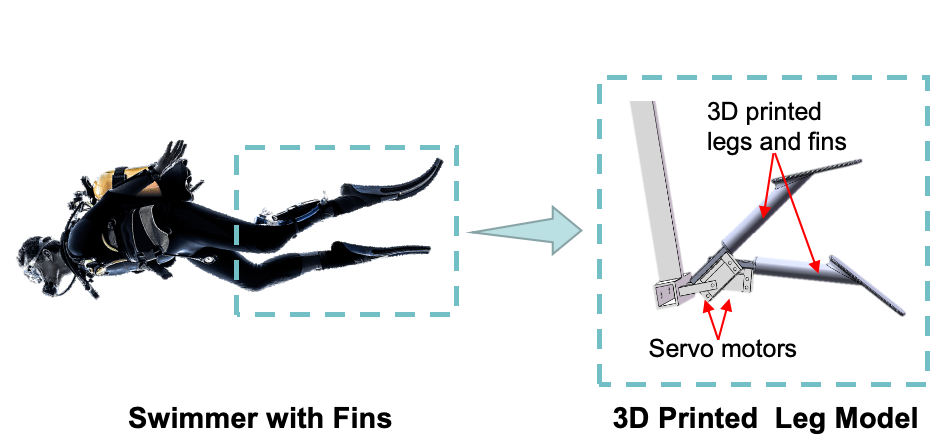}
    \caption{Illustration of the design of the scaled leg model mimicking human swimmers. The leg model is 3D printed, and actuated using two servo motors, simulating the kicking motion of swimmers wearing fins.
    }
    \label{fig:Simplify}
\end{figure}
This study investigates the use of the artificial lateral line (ALL) system to sense undulatory flow fields generated by human swimmer's leg kicks. Previous research shows a correlation between leg kick patterns and tempospatial features of the resulting undulatory flow fields\cite{Hochstein_2011_HMS,zamparo_2002_JEB}. Kick patterns are primarily characterized by the oscillation frequency and phase difference between the legs, resulting in two main styles, i.e., the dolphin kick and the flutter kick.

\begin{figure*}[!htb]
    \centering
    \subfigure[]{\includegraphics[width=0.45\linewidth]{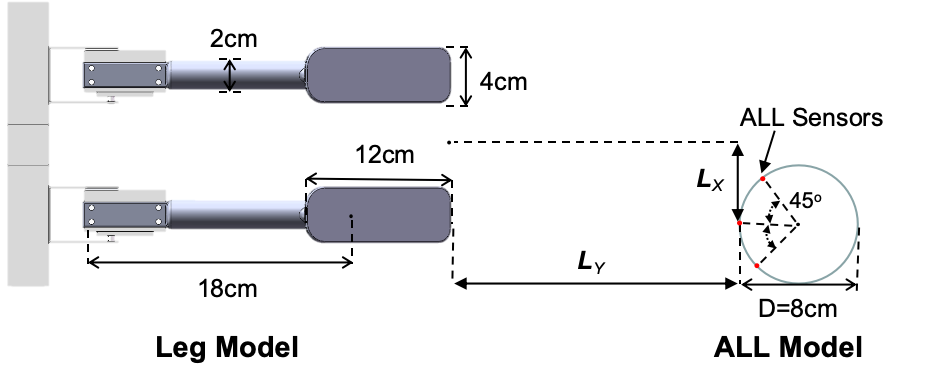}\label{fig: Q1}}
   \subfigure[]{\includegraphics[width=0.45\linewidth]{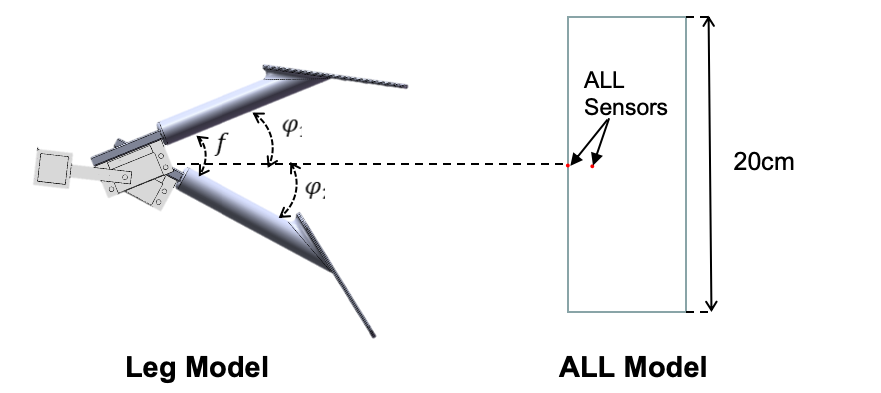}\label{fig: Q2}}
    \caption{Schematic diagram of the ALL-based flow sensing for the undulatory flow field generated by leg kicks. The ALL system samples flow pressure distribution of the downstream wake of swimmer's leg kicks for sensing tasks such as kick pattern recognition and kicking leg localization. (a) Top view illustrating the dimensions of the leg model and ALL system, and (b) side view illustrating the flutter kick pattern.}
    \label{fig: Problem}
\end{figure*}

This paper utilizes 3D printing to create a 1:5 scale model of swimmer's legs with fins, as illustrated in Fig.~\ref{fig:Simplify}. Two servo motors drive the leg model to simulate kicking motion. Following previous research\cite{zamparo_2002_JEB}, swimmer's leg kicks are designed to follow sine functions, i.e.,
\begin{equation}
    \begin{aligned}
        A_1(t) = A\sin(2\pi f t+ \varphi_1),  \\
        A_2(t) = A\sin(2\pi f t + \varphi_2)
        \end{aligned}
\end{equation}
where $A_1$ and $A_2$ are the amplitudes of motion of the left and right legs, respectively, $f$ is the kick frequency, and $\varphi_1$ and $\varphi_2$ are the initial phases of the left and right legs, respectively.
Two estimation tasks are particularly investigated in this paper.

\textbf{Task\,\#1: Leg Kick Pattern Recognition}
This paper classifies kick patterns by the kicking style (dolphin or flutter) and kicking frequency. To study the feasibility of using ALL to sense various kick patterns in practice, we establish a kick pattern set $\boldsymbol{S}=\{s_1,s_2,s_3,s_4,s_5,s_6\}$ that includes six kick patterns. Here, $s_1$, $s_3$ and $s_5$ represent the kick patterns with dolphin kicks, i.e., $|\varphi_1-\varphi_2| =0$ at kick frequency $f=1\,\text{Hz}$, $f=1.5\,\text{Hz}$ and $f=2\,\text{Hz}$, respectively, while $s_2$, $s_4$ and $s_6$ represent the kick patterns with flutter kicks, i.e., $|\varphi_1-\varphi_2| =\pi$ at kick frequency $f=1\,\text{Hz}$, $f=1.5\,\text{Hz}$ and $f=2\,\text{Hz}$, respectively.

Illustrated in Fig.~\ref{fig: Problem}, given the time series of distributed pressure measurements $\boldsymbol{P}(t)$, sampled by the ALL system from the flow field generated by leg kicks, the task of kick pattern recognition is to identify the pattern or mode $s \in \boldsymbol{S}$ of the leg kicks.

\textbf{Task\,\#2: Kicking Leg Localization}

In addition to identifying kick patterns, estimating the location of kicking legs plays an important role in HRI between human swimmers and underwater robots. This paper focuses on the two-dimensional kicking leg localization, specifically, the lateral and longitudinal displacements $L_x$ and $L_y$, as illustrated in Fig.~\ref{fig: Problem}. Given the time series $\boldsymbol{P}(t)$ of distributed pressure measurements, sampled by the ALL system from the flow field generated by leg kicks, the task of kicking leg localization is to find fitting functions $f_1$ and $f_2$ that estimate the lateral and longitudinal displacement $(L_x, L_y)$, i.e., 
\begin{align}
    \min_{f_1,f_2} \int \left\lVert (L_x,L_y) - (\hat{L}_x,\hat{L}_y) \right\lVert\mathrm{d}t.
\end{align}
Here, $\hat{L}_x = f_1(\boldsymbol{P}(t))$ and $\hat{L}_y = f_2(\boldsymbol{P}(t))$ represent the estimated lateral and longitudinal displacements.

\section{Flow Sensing for Swimmer Leg Kicks}
\subsection{Sensing Framework Design Using Feature Fusion}
\begin{figure*}
    \centering
    \includegraphics[width=0.9\textwidth]{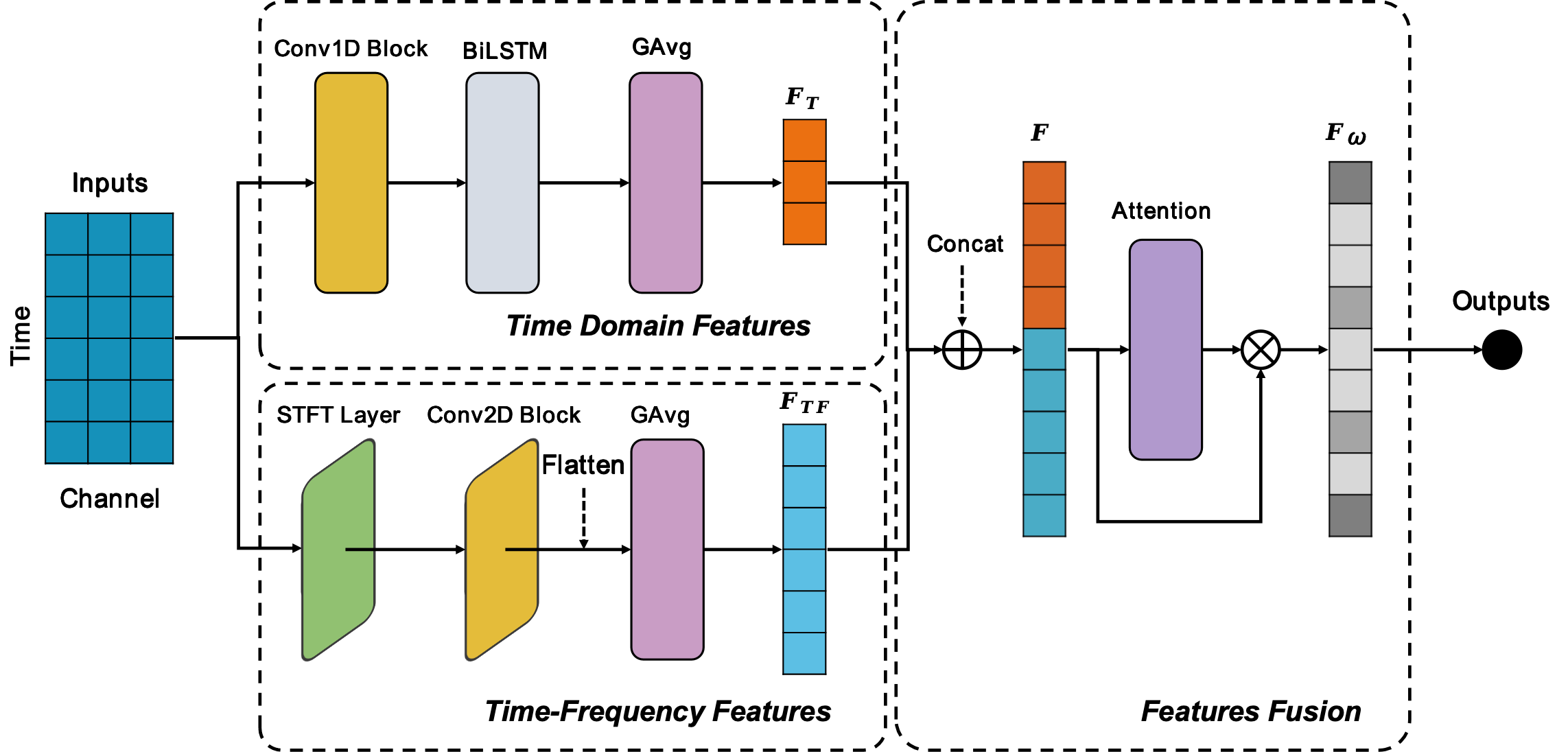}
    \caption{The proposed network architecture of the sensing framework for undulatory flow generated by swimmer's kicks. This framework consists of three primary modules, including the time-domain feature extraction module, the time-frequency domain feature extraction module, and the attention-based feature fusion module.}
    \label{fig:exp}
\end{figure*}

To address the tasks of kick pattern recognition and kicking leg localization, this paper proposes a novel flow-sensing framework that fuses time-domain and time-frequency domain features extracted from the temporospatial pressure measurements of the ALL system. By employing attention mechanisms, this sensing framework design facilitates adaptive feature fusion, ensuring accurate and robust flow sensing of swimmer leg kicks. 

The proposed sensing framework, illustrated in Fig.~\ref{fig:exp}, consists of three primary modules, including the time-domain feature extraction module, the time-frequency-domain feature extraction module, and the feature fusion module. The time series of distributed pressure measurements at time $t$ used for flow sensing, denoted as $\boldsymbol{P}(t)\in \mathscr{R}^{N_d \times N_s}$, where $N_d$ represents the number of the data points of the time series and $N_s$ represents the number of the ALL pressure sensors, is simultaneously inputted into both the time-domain and the time-frequency-domain feature extraction modules. 

In the time-domain feature extraction module, a one-dimensional convolutional neural network (1DCNN) learns local time-related features from the time series data $\boldsymbol{P}(t)$. These local features are then processed by a bidirectional long short-term memory (BiLSTM) network to learn feature representations. A global average pooling layer is applied to the BiLSTM output to reduce the feature dimensionality and enhance the model's generalizability.

For time-frequency-domain features, the short-time Fourier transform (STFT) converts the pressure time series data $\boldsymbol{P}(t)$ into a two-dimensional (2D) time-frequency representation. This representation is then processed by a two-dimensional convolutional neural network (2DCNN). A flatten layer converts the 2D time-frequency features into a one-dimensional (1D) feature vector. Similar to the time-domain processing, a global average pooling layer is applied to reduce the dimensionality of the time-frequency features and improve generalizability.
\begin{figure*}
    \centering
    \includegraphics[width=0.95\textwidth]{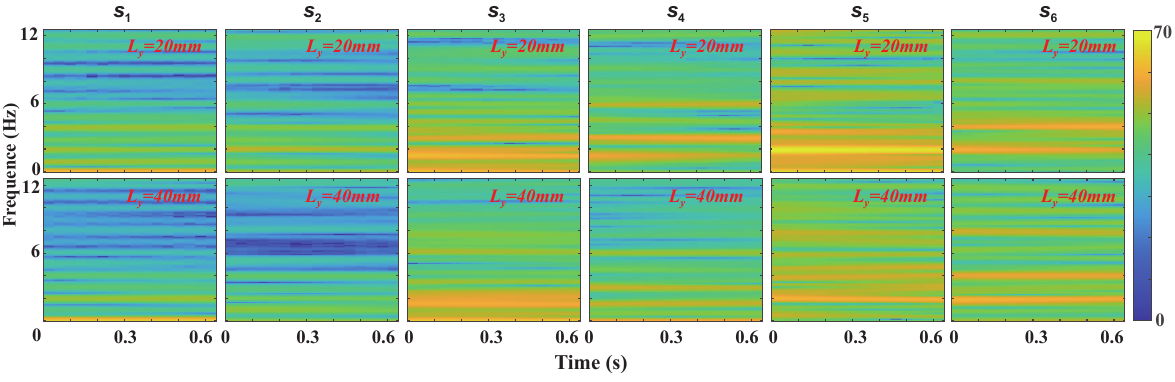}
    \caption{Illustration of the STFT spectrums of the distributed pressure measurements sampled by the ALL system in the undulatory flow field. The $x$-axis represents time, while the $y$-axis represents frequency. Each column shows the frequency spectrum for longitudinal displacements $L_y$ of 20\,mm or 40\,mm under the same leg kick pattern ${s}_i \in \boldsymbol{S}$, and each row shows the spectrum for different leg kick patterns ${s}_i$ at the same $L_y$.} 
    \label{fig:STFT}
\end{figure*}
To fully exploit the information from both time-domain and time-frequency-domain features, the time-domain vector $\boldsymbol{F}_\text{T}$ and time-frequency vector $\boldsymbol{F}_\text{TF}$ are concatenated into a single feature vector $\boldsymbol{F}$. This feature is passed through a dropout layer to prevent overfitting. Attention mechanisms dynamically adjust the weights of the features from the two domains. The resulting weighted feature vector $\boldsymbol{F}_\omega$ is then inputted into task-specific estimators to generate outputs of interest regarding the flow field of swimmer leg kicks. 

\subsection{Time-Frequency-Domain Feature Extraction Module}

The time-frequency feature extraction method has been crucial in traditional flow sensing for underwater oscillating objects \cite{ Qiu_2023_Tase}. In this paper, we extend this approach to undulatory flow fields generated by swimmer leg kicks, extracting time-frequency features from the pressure signals collected by the ALL system.

We employ the short-time Fourier transform (STFT), a prominent time-frequency analysis method, which segments a signal into short periods, each undergoing a Fourier transform to extract time-frequency features from distributed pressure measurements. Specifically, the kick frequencies of the leg models are set at 1\,Hz, 1.5\,Hz and 2\,Hz, respectively. To capture the kicking process comprehensively, data collected within a 4-second time window (100 data points) before the current time $n$ by the $i$-th pressure sensor is defined as the input data sequence $\boldsymbol{p}_i$ where $i=1,2,\ldots,N_s$. 
The time-frequency characteristics of these time series data are then analyzed using STFT with a Hamming window $\mathrm{w}(n)$ of an overlap length of 1\cite{Qiu_2023_Tase}. The STFT of the ALL pressure measurements is represented by
\begin{equation}
        X_i(m,k)=\sum_{n=0}^{N-1}\mathrm{w}(n)\boldsymbol{p}_i(n+mr)\exp{(-j2\pi n k/N)},
    \end{equation}
where $X_i(m,k)$ represents the amplitude of the $i$-th pressure sensor measurement at the $k$-th frequency component during the $m$-th time period, with $N$ being the number of points in the Fourier transform and $k$ a discrete frequency index ranging from $\left[0, N-1\right]$.
The spectrogram of the distributed pressure measurements is calculated as
\begin{equation}
        S_i(m,k)=|X_i(m,k)|^{2}.
\end{equation}

Figure~\ref{fig:STFT} illustrates the STFT spectrograms of the distributed pressure measurements under different leg kick patterns ${s}_i \in \boldsymbol{S}$, where $i=1,2,3,4,5,6$. Significant differences in frequency components are evident across different leg kick patterns. Additionally, variations in time-frequency characteristics within a single kick pattern ${s}_i$ are influenced by the relative displacement between the leg model and the ALL module. These observations highlight the importance of time-frequency domain features in sensing the undulatory flow field generated by swimmer leg kicks.

Given the proven effectiveness of CNNs in fields such as computer vision and speech recognition, particularly for 2D image processing, we construct a deep learning model that sequentially combines STFT and 2DCNN to extract time-frequency features from the distributed pressure measurements, thereby sensing the undulatory flow field.
    
\begin{figure*}[htp!]
        \centering
        \includegraphics[width=0.95\linewidth]{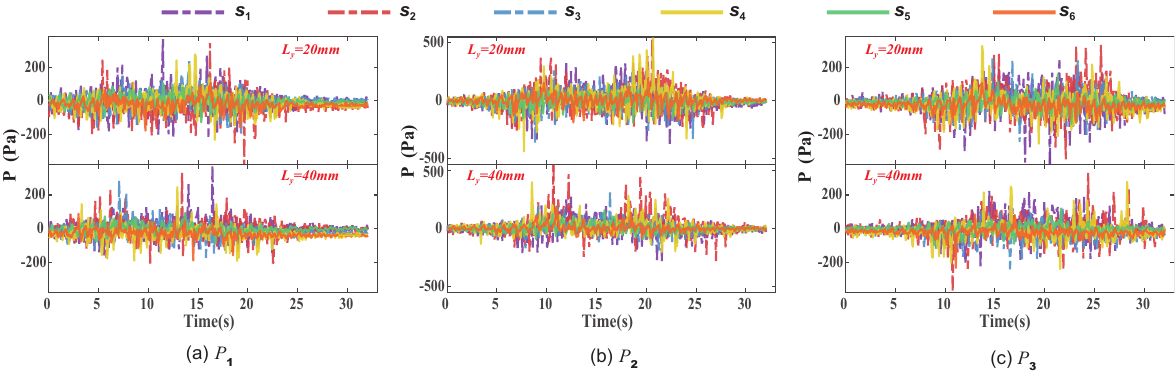}
        \caption{
        Illustration of distributed pressure measurements collected by three pressure sensors ($P_1, P_2, P_3$) in the ALL system. Subfigures~5(a), 5(b), and 5(c) show the data from sensors $P_1$, $P_2$, and $P_3$, respectively. Different colored curves represent various leg kick patterns ${s}_i \in \boldsymbol{S}$: purple dashed line for ${s}_1$,  red dashed line for ${s}_2$, blue dashed line for ${s}_3$, yellow line for ${s}_4$, green line for ${s}_5$ and orange line for ${s}_6$. Each figure contains a top subfigure showing pressure data at a longitudinal displacement $L_y$ of 20\,mm and a bottom subfigure showing data at $L_y$ of 40\,mm.
        }
        \label{fig:sensor_reading}
\end{figure*}
\subsection{Time-Domain Feature Extraction Module}

While time-frequency features excel in discerning the relationship between the time and frequency of pressure signals, they often fall short in capturing the temporal relationship between sequential data points. In contrast, time-domain features encapsulate these temporal relationships effectively, which is crucial in sensing the undulatory flow field~\cite{Xu_2022_BIB,Xu_2023_BIB}. Fig.~\ref{fig:sensor_reading} illustrates pressure sequences collected from three sensors ($P_1$, $P_2$, and $P_3$) under different leg kick patterns ${s}_i \in \boldsymbol{S}$ where $i=1,2,3,4,5,6$. Notable disparities in the time domain across different kick patterns indicate the need for a time-domain feature extraction module.

The effectiveness of 1DCNN and LSTM networks in time-domain feature extraction is well-established, as examplified by our prior work in propeller wake sensing~\cite{Wang_2024_Tmech}. This study adopts a similar 1DCNN and BiLSTM hybrid model to extract time-domain features.
Specifically, the 1DCNN captures local features with temporal dependencies from the time series of pressure data. These local features are then inputted into the BiLSTM network to encode the temporal dependencies further. The BiLSTM processes the ALL pressure time sequence data $\boldsymbol{P}(t)$ to derive global time-domain features. While LSTM networks are proficient in learning global features from sequence data, BiLSTM enhances this capability by learning both forward and backward features of the time sequence. This bidirectional learning enables the extraction of more comprehensive global features, thus improving the learning of long-term dependencies.

\subsection{Attention-Based Dynamic Feature Fusion}
To leverage both time-domain features $\boldsymbol{F}_\text{T}$ and time-frequency-domain features $\boldsymbol{F}_\text{TF}$ extracted from the pressure time series data $\boldsymbol{P}=[\boldsymbol{p}_1^T, \boldsymbol{p}_2^T,\boldsymbol{p}_3^T]^T$, we combine these feature vectors and obtain the overall feature representation $\boldsymbol{F}$, i.e., $\boldsymbol{F}=\boldsymbol{F}_\text{T}\oplus \boldsymbol{F}_\text{TF}$ where $\oplus$ denotes the concatenation operator.

An attention mechanism\cite{NIPS2017_Vaswani} is then adopted to optimize the relative contributions of the extracted features, dynamically adjusting the importance weights $\boldsymbol{\omega} \in \mathcal{R}^{1 \times N}$ and biases $\boldsymbol{\beta}\in \mathcal{R}^{1 \times N}$, where $N$ denotes the total number of the features $\boldsymbol{F}$, enabling the model to focus on the most critical aspects of the tempospatial features.
Specifically, the self-attention mechanism designed in this paper computes the weight vector $\boldsymbol{\omega}$ by
\begin{equation}
    \boldsymbol{\omega} = \operatorname{softmax}\left(\boldsymbol{\omega}^T \boldsymbol{F}+\boldsymbol{\beta}\right),
\end{equation}
where $\mathrm{softtmax}(\cdot)$ represents the softmax activation function.
The weights $\boldsymbol{\omega}$ are assigned to the extracted flow features $\boldsymbol{F}$ through element-wise multiplication, obtaining weighted feature representation $\boldsymbol{F}_\omega$, i.e.,
\begin{equation}
    \boldsymbol{F}_\omega=[\boldsymbol{F}_1, \boldsymbol{F}_2,\ldots,\boldsymbol{F}_N] \otimes [{\omega}_1, {\omega}_2, \ldots, {\omega}_N],
\end{equation}
where $\boldsymbol{F}_i$ denotes the $i$-th feature of unweighted $\boldsymbol{F}$, $w_i$ denotes the weight corresponding to the $i$-th feature, $\otimes$ represents the element-wise multiplication operator, and $\boldsymbol{F}_\omega$ represents the final fusion of time-domain and time-frequency features of the undulatory flow field.
Combining time-domain and time-frequency features through concatenation and adaptive weighting via an attention mechanism allows the model to effectively utilize complementary information, enhancing the performance and robustness of the sensing task.

\begin{figure*}[htp!]
    \centering
    \includegraphics[width=0.9\linewidth]{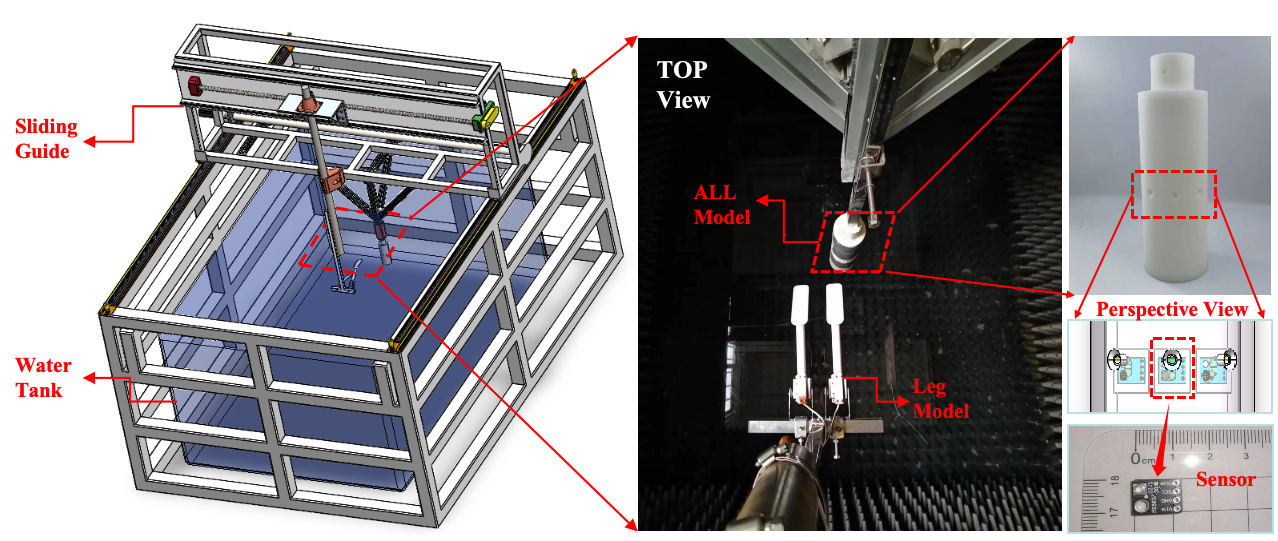}
    \caption{Illustration of the experimental platform design and the bioinspired ALL system. The testing platform consists of a pool and a high-precision sliding guide. The pool dimensions are 3\,m~(L) $\times$ 2\,m~(W) $\times$ 1.5\,m~(D). A 1:5 scaled human leg model with fins is developed through 3D printing. The ALL system follows a cylindrical shape, equipped with three distributed MS5837-02Ba sensors for measuring their local pressures.}
    \label{fig:platform}
\end{figure*}

\section{Experiments}
\subsection{Experimental Platform}

This study utilizes the ALL system to investigate the undulatory flow field generated by swimmer's leg kicks. Section II describes constructing a 1:5 scale human leg model with fins through 3D printing, as illustrated in Fig.~\ref{fig: Problem}. The leg has a diameter of 2\,cm and a length of 18\,cm. The fins measure 3\,cm in width and 12\,cm in length with a fin angle of 60$^\circ$. The leg model is driven by two servo mechanisms, simulating swimmer's kicks and generating an undulatory flow field.

In ALL-based sensing, circular shapes are widely adopted for their analytical models of fluid dynamics. The ALL system in this study adopts a cylindrical shape, as illustrated in Fig.~\ref{fig:platform}, with a radius of 4\,cm and a height of 20\,cm. To enable distributed pressure measurements, three pressure sensing holes are uniformly arranged with 45$^\circ$ angular separation and 10 cm above the bottom. These holes have a diameter of 4\,mm and are connected to commercial pressure sensors via silicone tubing. The MS5837-02Ba pressure sensor from TE Connectivity is selected, utilizing I2C bus communication to transmit pressure data to the host computer at a sampling rate of 25\,Hz, with a pressure resolution of 1\,Pa.

As illustrated in Fig.~\ref{fig:platform}, a testing platform is constructed consisting of a pool and a high-precision sliding guide. The pool dimensions are 3\,m~(L) $\times$ 2\,m~(W) $\times$ 1.5\,m~(D). The high-precision sliding guide provides support and precise motion capability for the system. The ALL system is fixed on the support frame of the pool, and the leg model is rigidly fastened to the high-precision sliding guide, controlled by an industrial computer for movement. 

\subsection{Experimental Configurations}
This paper primarily aims to sense the undulatory flow field generated by swimmer's leg kicks by utilizing the distributed pressure measurements $\boldsymbol{P}(t)$ from the ALL system. 
In this study, the motion of the leg model is selected as lateral movements at a constant speed (500\,mm/min) with various constant longitudinal displacements. Consequently, $L_x$ is a continuous value, while $L_y$ is discrete. 

To assess the ALL system's flow field sensing ability within $1\,\mathrm{m}^2$ behind swimmers in practical scenarios, considering the 1:5 scale of the leg model, the lateral displacement $L_x$ takes a range of 200\,mm in the experiment, i.e., $L_x \in \left[-100,100\right]$\,mm, and the longitudinal displacement $L_y$ is distributed at intervals of 20\,mm ranging from 20\,mm to 200\,mm. 
Additionally, based on the leg model settings in Section II, this experiment identifies four kick patterns within $\boldsymbol{S}=\left\{s_1,s_2,s_3,s_4,s_5,s_6 \right\}$, combining different kick styles and frequencies. Experimental configurations are detailed in Table~I.

\begin{table}[]
    \centering
    \caption{Experimental configurations}
    \label{tbl:parameters}
    \begin{tabular}{ccc}
    \Xhline{2pt}
    \multicolumn{2}{c}{State} & Parameters \\ \Xhline{1pt}
    \multirow{4}{*}{$\boldsymbol{S}$}   & $s_1$   &  $f=1$ Hz,  $\left | \varphi_1-\varphi_2  \right | =\pi $      \\ \cline{2-3} 
                         &  $s_2$  &  $f=1$ Hz,  $\left | \varphi_1-\varphi_2  \right | =0 $          \\ 
                         \cline{2-3} 
                         &  $s_3$  &  $f=1.5$ Hz,  $\left | \varphi_1-\varphi_2  \right | =\pi $          \\ \cline{2-3} 
                         &  $s_4$  &  $f=1.5$ Hz,  $\left | \varphi_1-\varphi_2  \right | =0 $         \\
                         \cline{2-3} 
                         &  $s_5$  &  $f=2$ Hz,  $\left | \varphi_1-\varphi_2  \right | =\pi $          \\ \cline{2-3} 
                         &  $s_6$  &  $f=2$ Hz,  $\left | \varphi_1-\varphi_2  \right | =0 $         \\ \Xhline{1pt}
                         \multicolumn{2}{c}{$L_x$}   &   $\left [-100,100\right ]$\,mm         \\ \Xhline{1pt}
                         \multicolumn{2}{c}{$L_y$}   &   $\left\{ 20,40,60,80,100,120,140,160,180,200 \right\}$\,mm         \\ \Xhline{2pt}
\end{tabular}
\end{table}

\begin{figure*}[!htb]
    \centering
    \subfigure[ The proposed feature fusion.]{\includegraphics[width=0.32\linewidth]{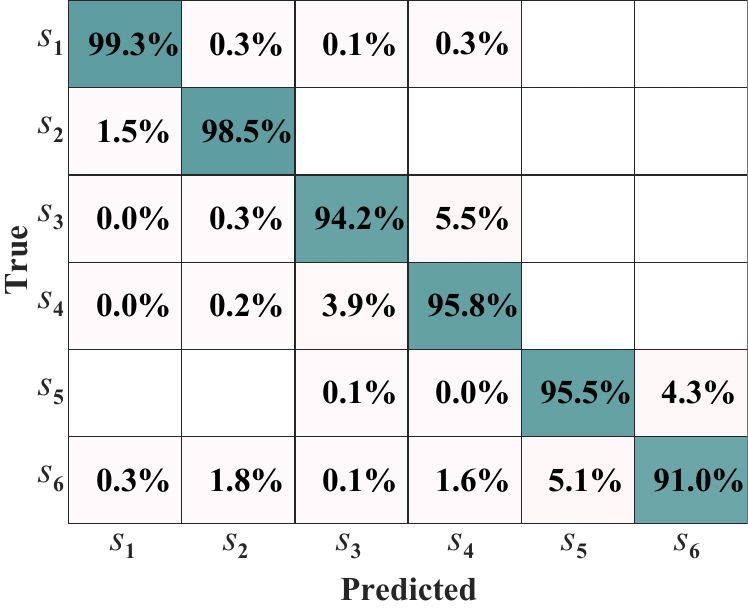}\label{fig: Hybird}}
    \subfigure[Time-domain-based 1DCNN-BiLSTM.]{\includegraphics[width=0.32\linewidth]{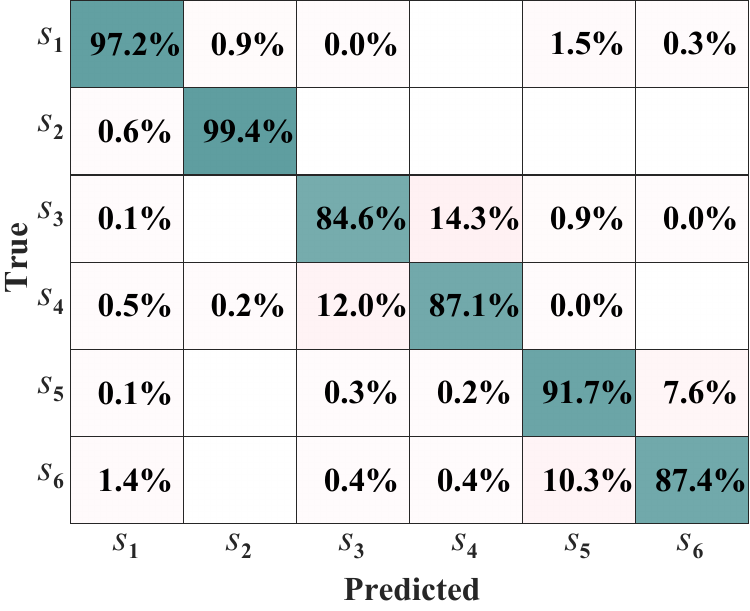}\label{fig: Time8755}}
    \subfigure[Time-frequency-based STFT-2DCNN.]{\includegraphics[width=0.32\linewidth]{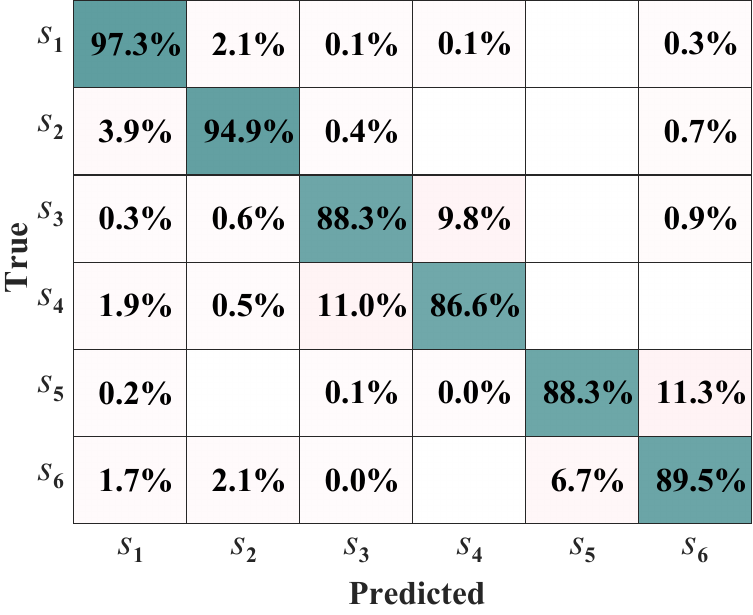}\label{fig: STFT8308}}
   
    \caption{The experimental results of ALL-based leg kick pattern recognition along with ablation study results on the feature extraction module comparing the sensing performance between the proposed feature fusion method, the time-domain-based 1DCNN-BiLSTM method, and the time-frequency-based STFT-2DCNN method. In the confusion matrix, the $y$-axis represents the true values, the $x$-axis represents the predicted values, and the green shade indicating the accuracy level. The proposed sensing method achieves satisfactory identification of different leg kick patterns.} 
    \label{fig: Motion State}
\end{figure*} 
\subsection{Data Acquisition and Processing}
To construct a dataset for traning and testing the proposed sensing algorithm, extensive experiments are conducted on the experimental platform developed in this paper. Prior to the motion of the leg model, distributed pressure measurements are collected as reference pressures, representing the sum of atmospheric pressure and static water pressure. During each experiment, the legs move back and forth along $L_x$ at a constant speed in pattern $s_i\in\boldsymbol{S}$, while the longitudinal displacement $L_y$ remains constant. Considering the effective measurement width of the ALL system on the leg kick flow and the transient effects during the leg's lateral motion, the leg model moves from $L_x=-175$\,mm to $L_x=175$\,mm and then returns to $L_x=-175$\,mm. The effective estimation range is clipped to $L_x \in \left [-100,100\right ] $\,mm, with multiple experiments conducted to traverse the leg kick patterns $s_i\in\boldsymbol{S}$ and longitudinal displacements $L_y$. Each set of experimental configurations in Table I is repeatedly tested ten times. 

To eliminate the influence of environmental noises, all pressure measurements used in flow sensing are adjusted by subtracting the mean steady-state pressure computed from the initial pressure samples collected when the leg model is at rest, forming the input data to the sensing network $\boldsymbol{p}_{i}(t)$ for the $i$-th pressure sensor. The corresponding leg displacements and kick patterns are labeled, along with the input data $\boldsymbol{P}(t)$, establishing a flow-sensing dataset for swimmer's leg kicks (The Dataset is available online). 

\subsection{Training Environment and Hyperparameters}

The training is conducted on a desktop computer with an Intel Core i9-129000K CPU, 64BG RAM, and a GeForce RTX 3080Ti GPU, using MATLAB 2023b. While sharing feature extraction modules, the kick pattern recognition task adopts the classification output design while the kicking leg localization adopts the regression method, leading to variations in hyperparameters.
For kick pattern recognition, the training steps are set to 30 with a batch size of 128. The learning rate, initially set as 0.005, gradually decreases as training proceeds. In kicking leg localization, the training steps are increased to 1000, with the same batch size of 128 and learning rate settings.

\section{Experimental Results and Analysis}

\subsection{Kick Pattern Recognition Task}

\begin{table}
    \centering
    \label{tbl:table1}
    \caption{Average classification accuracy of kick pattern recognition}
    \begin{tabular}{cc}
        \Xhline{2pt}
        
         {Methods} &{Accuracy}\\
        \Xhline{1pt}
        
   1DCNN-BiLSTM &  91.24\%\\
 STFT-2DCNN& 90.79\%\\
     Our Proposed   & \textbf{95.69\%}\\
        \Xhline{2pt}
    \end{tabular} 
\end{table}
To validate the effectiveness of the proposed method in sensing the undulatory flow field generated by swimmer's leg kicks, we experimentally test kick pattern recognition. Six kick patterns are considered in this paper as detailed in Table I.

Figure~\ref{fig: Motion State}(a) presents the experimental results of the leg kick pattern recognition, illustrated as a confusion matrix, with the $y$-axis representing the true values, the $x$-axis representing the predicted values, and the green shade indicating the accuracy level. We observe that the sensing accuracy is exceeding 90\% across all the six kicking patterns, suggesting that the proposed sensing method achieved accurate identification of various leg kick patterns. 

Figures~\ref{fig: Motion State}(b) and \ref{fig: Motion State}(c) present the ablation study results of the proposed sensing framework, comparing the sensing performances between using time-domain-only (1DCNN-BiLSTM) and time-frequency-domain-only (STFT-2DCNN) features. 
The proposed feature fusion method significantly enhances kick pattern recognition, with estimation accuracy improving across almost all patterns. Notably, the lowest estimation accuracy of our proposed method is 91\%, compared to 84.6\% and 86.6\% from time-domain and time-frequency-domain features, respectively.
\begin{figure}
    \centering
    \includegraphics[width=0.951\linewidth]{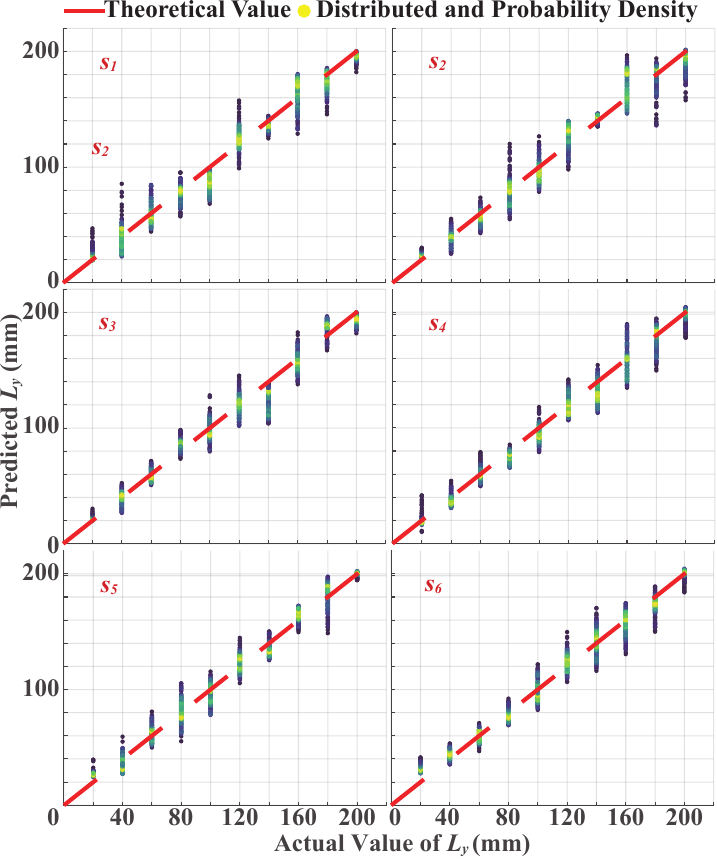}
    \caption{The experimental results of ALL-based longitudinal displacement $L_y$ sensing under various leg kick patterns. Colored dots represent estimated values ($y$-axis) versus actual values ($x$-axis). Point colors indicate probability density with yellower for higher density and bluer for lower density. The red dashed line represents perfect estimation. 
    }
    \label{fig:Longitudinal_displacement}
\end{figure}
Table II summarizes the ablation study results, comparing the averaged accuracy of the classification results achieved by different feature extraction methods for sensing leg kick patterns. 
From the comparison results, we observe that the feature fusion design surpasses single-domain feature methods, improving estimation accuracy by approximately 4\% over the time-domain method and approximately 5\% over the frequency-domain method, demonstrating superior performance in leg kick pattern sensing.

\subsection{Kicking Leg Localization Task}

\begin{figure}[htp!]
    \centering
    \includegraphics[width=0.98\linewidth]{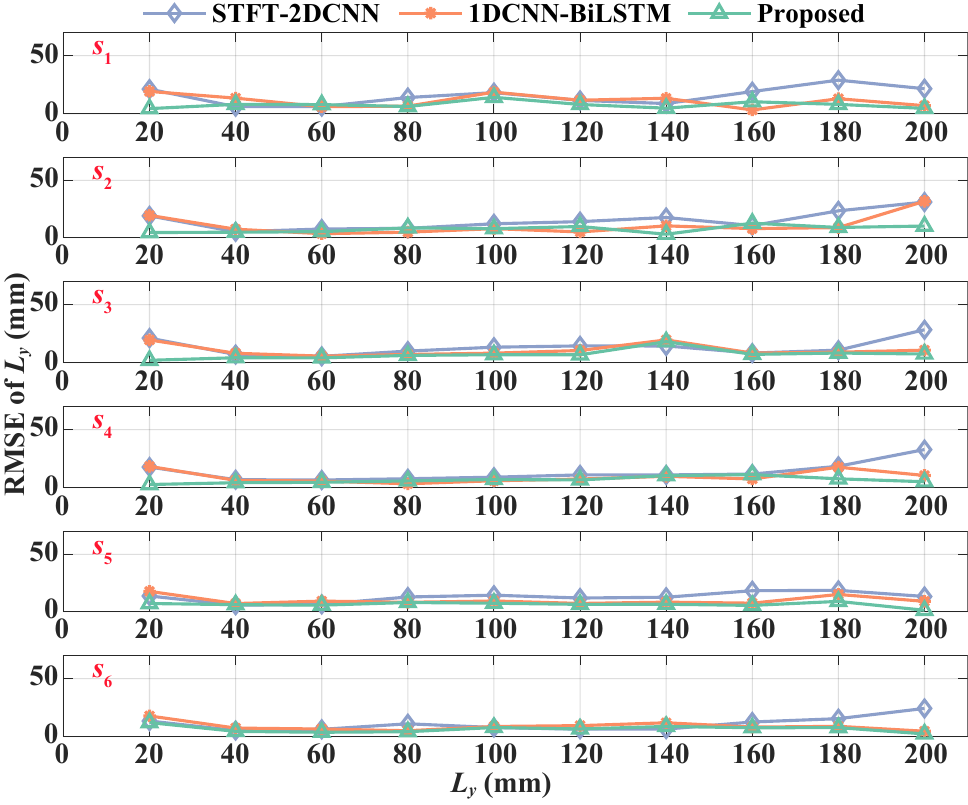}
    \caption{The experimental results of RMSE for longitudinal displacement sensing along with  ablation study results on the feature extraction module comparing the sensing performance between the proposed feature fusion method, the time-domain-based 1DCNN-BiLSTM method, and the time-frequency-based STFT-2DCNN method under different leg kick patterns.}
    \label{fig:y_dis}
\end{figure}

In the kicking leg localization task, we examine the proposed sensing method in estimating the longitudinal and lateral displacements of the leg model measured from the ALL sensing system.

\subsubsection{Longitudinal Displacement Sensing}

\begin{table}[]
\centering
    \label{tbl:table2}
    \caption{Longitudinal displacement $L_y$ sensing results}
\begin{tabular}{ccccccc}
\Xhline{2pt}
\multirow{2}{*}{Methods} & \multicolumn{6}{c}{RMSE of $\hat{L}_y$} \\ 
\cline{2-7} 
                          & $s_1$ & $s_2$ & $s_3$ & $s_4$ & $s_5$ & $s_6$ \\  
\Xhline{1pt}
 1DCNN-BiLSTM             & 14.8 & 14.9 & 13.1 & 11.8& 12.4& 11.7\\
STFT-2DCNN                &19.1   & 18.6  & 17.0  & 17.1&15.8&13.9\\
 Our Proposed            &\textbf{10.1} & \textbf{10.3} & \textbf{9.5} & \textbf{8.9} &\textbf{7.9} & \textbf{8.5}\\ \Xhline{2pt}
\end{tabular}
\end{table}

The experimental results of the longitudinal localization are illustrated in Fig.~\ref{fig:Longitudinal_displacement}, which presents the estimates of the longitudinal displacement $L_y$ between the leg model and the ALL model. Ten discrete displacement values are used as described in Table~I. 
The subfigures display the estimated displacement $\hat{L}_y$ for different leg kick patterns $s_i\in\boldsymbol{S}$ where $i=1,2,3,4,5,6$. 
The colored dots represent the estimated values, with the $y$-axis for estimated results and the $x$-axis for actual values. The color of each point indicates the probability density, where yellower suggests higher density and bluer suggests lower density. The red dashed line indicates the perfect estimated values. 

The results show that the estimates are clustered around the perfect estimate values with approximately 90\% accuracy (within 2\,cm), though outliers occur, particularly in $s_1$ and $s_2$, with deviations up to 6\,cm but a probability less than 10\%. Kick patterns $s_5$ and $s_6$ demonstrate better accuracy compared to $s_1$ and $s_2$, likely due to higher leg kick frequencies that increase the flow velocity, thereby enhancing the sensor's Signal-to-Noise Ratio (SNR).

\begin{figure*}[htp!]
    \centering
    \includegraphics[width=0.95\linewidth]{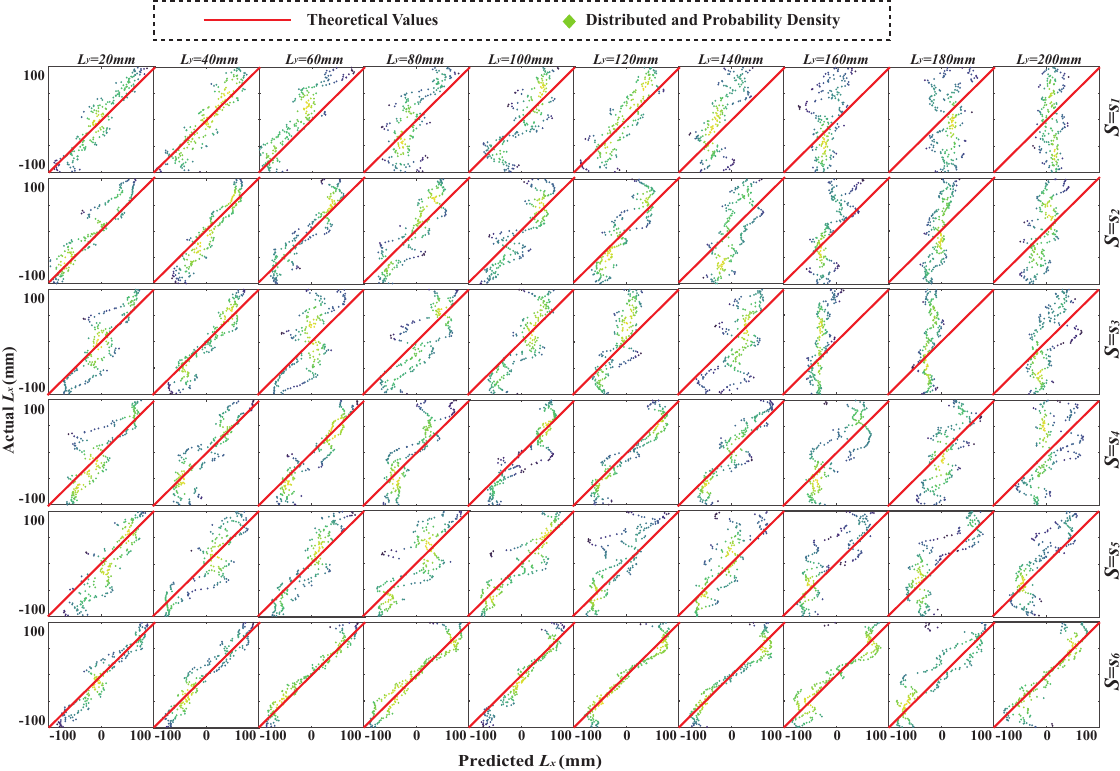}
    \caption{The experimental results of ALL-based lateral displacement $L_x$ sensing under various leg kick patterns and longitudinal displacements $L_y$. The rows denote different leg kick patterns $s_i$, and the columns represent different longitudinal displacements $L_y$. The $y$-axis shows the actual values, while the $x$-axis shows the estimated values. A red line indicates the perfect estimates. The color gradient of the points represents the probability density, with yellower indicating higher density.}
    \label{fig:x_dis_result}
\end{figure*}

Figure~\ref{fig:y_dis} and Table III illustrate the ablation study results comparing the root mean square error (RMSE) for estimated longitudinal displacement $\hat{L}_y$ between using single-domain features and the proposed feature fusion method across all leg kick patterns $s_i\in\boldsymbol{S}$ where $i=1,2,3,4,5,6$.  Each point represents the RMSE, with the green point indicating the proposed feature fusion-based sensing method. This method significantly reduces estimation errors, showing a 30.40\% and 46.07\% decrease in maximum RMSE across all leg kick patterns compared to using time-domain features only (1DCNN-BiLSTM) and time-frequency-domain features only (STFT-2DCNN), respectively.

The comparison results demonstrate that the localization error of the proposed method consistently stays below 20\,mm across various leg kick patterns. Given a sensing distance of 200\,mm, this  error is less than 10\%, suggesting robust underwater localization performance for potential applications.
Additionally, similar to the observations in kick pattern classification experiments, the experimental results indicate that higher leg kick frequencies result in lower sensing errors. This improvement is likely due to increased water flow velocity at higher frequencies, which enhances the robustness of pressure measurements and improves associated SNR.

\subsubsection{Lateral Displacement Sensing}

Figure~\ref{fig:x_dis_result} illustrates the results of lateral localization, showing the estimated relative lateral displacement $\hat{L}_x$ between the leg model and the ALL system. According to the experimental configuration (outlined in Table I), $L_x$ is defined over a continuous interval, representing various lateral displacements evaluated at selected discrete longitudinal displacement $L_y$ under different kick patterns $s_i\in\boldsymbol{S}$ where $i=1,2,3,4,5,6$. 

In Fig.~\ref{fig:x_dis_result}, the results are presented in a grid format. The rows denote different leg kick patterns $s_i$, and the columns represent different longitudinal displacements $L_y$. The $y$-axis shows the actual values, while the $x$-axis shows the estimated values. A red line indicates the perfect estimates. The color gradient of the points represents the probability density, with yellower indicating higher density.
The experimental results reveal that the proposed method achieves accurate lateral displacement $L_x$ estimates when longitudinal displacement $L_y$ is less than 100\,mm. These estimated values are closely aligned with the perfect estimate red line. However, beyond 100\,mm longitudinal displacement, the accuracy of the estimates declines as the estimated values increasingly deviate from the actual values.
\begin{figure}[htp!]
    \centering
    \includegraphics[width=\linewidth]{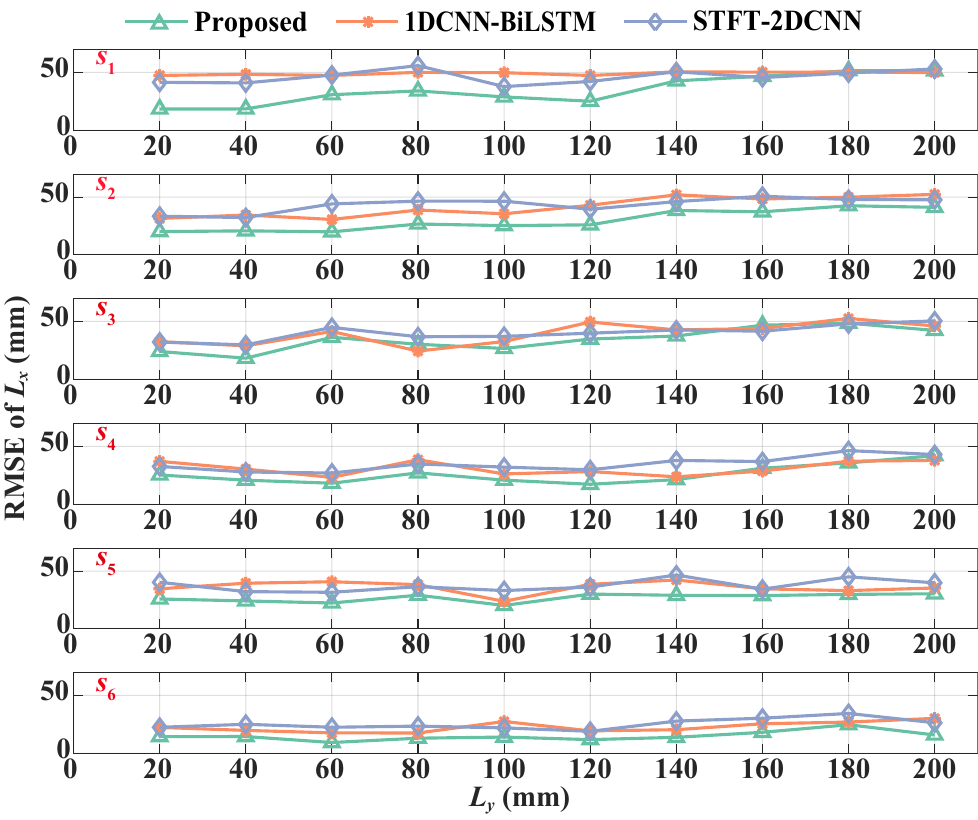}
     \caption{The experimental results of RMSE for lateral displacement sensing along with ablation study results on the feature extraction module comparing the sensing performance between the proposed feature fusion method, the time-domain-based 1DCNN-BiLSTM method, and the time-frequency-based STFT-2DCNN method under different leg kick patterns.}
    \label{fig:x_dist}
\end{figure}
\begin{table}[]
\centering
\label{tbl:table3}
\caption{Lateral displacement $L_x$ sensing results}
\begin{tabular}{ccccccc}
\Xhline{2pt}
\multirow{2}{*}{Methods} & \multicolumn{6}{c}{RMSE of $\hat{L}_x$} \\ 
\cline{2-7} 
                          & $s_1$ & $s_2$ & $s_3$ & $s_4$ & $s_5$ & $s_6$ \\  
\Xhline{1pt}
1DCNN-BiLSTM             & 56.7 & 50.8 & 50.0 & 39.6 & 42.9 & 31.1 \\
STFT-2DCNN               & 55.5 & 53.9 & 48.5 & 44.2 & 45.5 & 32.3 \\
\textbf{Our Proposed}     & \textbf{43.5} & \textbf{36.5} & \textbf{42.7} & \textbf{33} & \textbf{33.7} & \textbf{20.9} \\ 
\Xhline{2pt}
\end{tabular}
\end{table}

Figure~\ref{fig:x_dist} and Table IV illustrate the ablation study results on the proposed sensing network, comparing the root mean square error (RMSE) of estimated displacement $\hat{L}_y$ for various longitudinal displacement $L_y$ under different leg kick patterns $s_i\in\boldsymbol{S}$ where $i=1,2,3,4,5,6$, with the green line representing the proposed sensing method in this study. 
Compared to time-domain (1DCNN-BiLSTM) and time-frequency-domain (STFT-2DCNN) feature extraction methods, the proposed method achieves lower estimation errors, with a significant reduction in the maximum RMSE of about 23\% compared to 1DCNN-BiLSTM and 21\% compared to STFT-2DCNN.

Additionally, we observe from the experimental results that for $L_y$ less than 100\,mm, the proposed method maintains a lateral localization error of less than 40\,mm across different leg kick patterns, representing less than 20\% of the 200\,mm sensing distance and demonstrating its effectiveness in underwater localization. However, as longitudinal displacement $L_y$ exceeds 100\,mm, the localization accuracy for $L_x$ decreases. Furthermore, the experimental results also shows that leg kick frequency significantly affects localization accuracy, with lower sensing errors observed at higher frequencies, consistent with longitudinal localization results.

\subsection{Influence of Displacement on Localization}

In undulatory flow fields generated by leg kicks, flow vortices dissipate due to fluid viscosity, reducing energy, flow velocity, and pressure variations measured by the ALL system. This reduction leads to a decreased SNR, impacting localization accuracy. This section examines how displacement affects localization.

Regarding longitudinal displacement sensing, the proposed feature fusion method outperforms time-domain and time-frequency-domain methods. For instance, at $L_y=200$\,mm, the average RMSE for all leg kick patterns is approximately 51\% lower than that of the time-domain 1DCNN-BiLSTM method and 63\% lower than the time-frequency-domain STFT-2DCNN method. However, for $L_y\ge 200$\,mm, estimation errors increase, likely due to exceeding the ALL system's effective sensing range, resulting in reduced pressure values and decreased SNR. 

Figure~\ref{fig:x_dist} illustrates that, for lateral localization, the proposed feature fusion method surpasses the time-domain and time-frequency-domain methods, though distance significantly impacts longitudinal displacement sensing. Specifically, at $L_y=200$\,mm,  the average RMSE for all kick patterns is approximately 10\% lower than the time-domain 1DCNN-BiLSTM and 15\% lower than the time-frequency-domain STFT-2DCNN methods. 
As $L_y$ increases, the RMSE for lateral displacement estimates $\hat{L}_x$ also increases, with an average increase of 130\% from $L_y=20$\,mm to $L_y=200$\,mm across all kick patterns. We conjecture that this trend is due to the gradual decrease in energy and flow velocity with increased propagation distance, reducing pressure changes measured by the ALL system and decreasing SNR.

In summary, the distance between the ALL sensing system and the leg kicks affects localization accuracy and limits practical applications. However, the proposed attention-based feature fusion method enhances the robustness of flow sensing as distance increases.

\subsection{Comparison With Traditional Flow Sensing Methods}
\begin{figure}[htp!]
    \centering
    \includegraphics[width=0.95\linewidth]{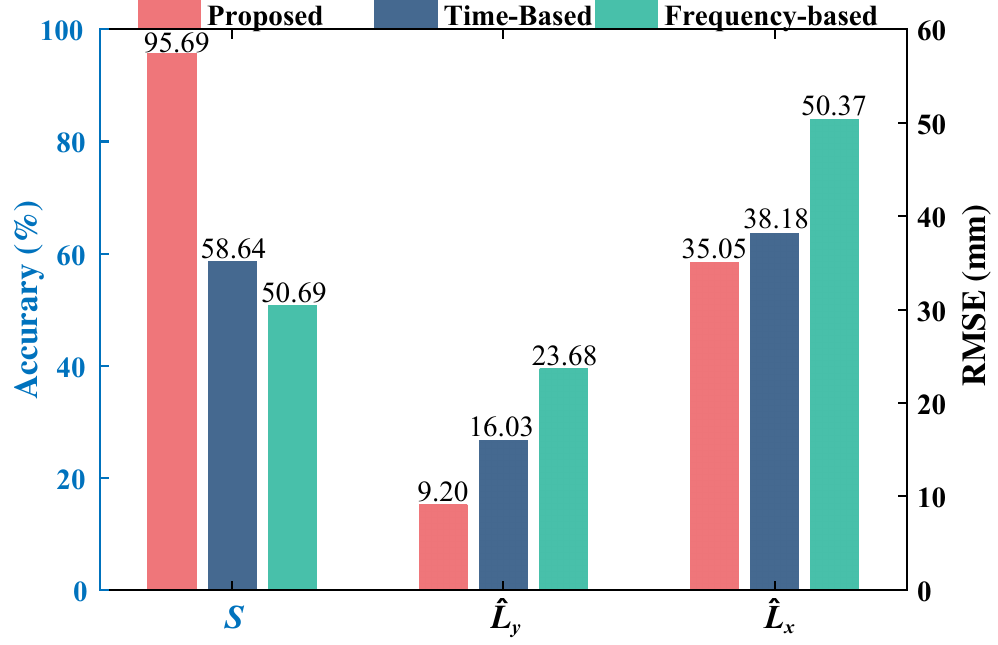}
     \caption{The comparative experimental results of leg kick pattern recognition and kicking leg localization between the proposed method and traditional frequency-domain-based and time-domain-based methods. Both traditional methods extract features without using neural networks. The left $y$-axis represents the accuracy of leg kick pattern ($S$) recognition, while the right $y$-axis represents the RMSE of lateral ($\hat{L}_x$) and longitudinal ($\hat{L}_y$) displacement estimates.}
    \label{fig:compare}
\end{figure}

To demonstrate the effectiveness of our proposed sensing method, we conducted a comparative experiment with two traditional flow sensing methods from the literature --- the frequency-domain-based method\cite{Guo_2023_CBS,Jiang_2022_Tmech} and the time-domain-based method\cite{Liu_2023_FMI,Venturelli_2012_BiB}. Both methods extract features without using neural networks. The frequency-based method employs the fast Fourier transform (FFT) to extract frequency features, specifically the first two frequency components of maximum energy and their amplitudes  extracted from the time series of pressure measurements. These features are then inputted into a multi-layer perceptron (MLP) to generate estimated states. The time-based method extracts statistical features from the time series, including the maximum, minimum, and mean values, and uses an MLP to obtain estimation results.

Figure~\ref{fig:compare} shows the comparison results of the statistical estimation accuracy regarding the tasks of interest. The comparison reveals that our proposed method consistently outperforms the conventional flow sensing algorithms across different tasks. Specifically, the average RMSE of the estimated longitudinal displacement $\hat{L}_y$ over all leg kick patterns decreases by 42.61\% and 61.15\%, and the average RMSE of the estimated lateral displacement $\hat{L}_x$ over all leg kick patterns decreases by 7.86\% and {30.41}\%, compared to time-based and frequency-based methods, respectively. Similarly, the accuracy of the leg kick pattern recognition increases by 37.05\% and 45\%, compared to the conventional time-based and frequency-based methods, respectively. These results illustrate the significant advantages of the proposed method over conventional flow sensing approaches in addressing the unique challenge of sensing the undulatory flow field generated by human leg kicks using the ALL system.
\section{Discussion}
\subsection{Limitations}
We would like to briefly discuss the limitations of the proposed undulatory flow filed sensing method and explore mitigation measures to enhance its real-world application.

First, the real-time performance of the algorithm is affected during pattern transitions, necessitating the development of multi-resolution recognition algorithms to process sensor data of varying time series lengths and prioritize newly updated data for improved accuracy and responsiveness. Second, the information space for HRI is currently limited. Expanding this space requires incorporating different kick amplitudes and enhancing the method's ability to sense these amplitudes through both algorithmic improvements and hardware optimizations. Third, the ALL system was fixed on the experimental platform, and the study did not account for sensor noise or pressure perturbations caused by its motion. Future work should integrate inertial measurement units (IMUs) and employ sensor fusion techniques to mitigate these effects and improve system robustness.

\subsection{The Potential for HRI between Underwater Robots and Human Swimmers}
The unique physical properties of water significantly influence the interaction between underwater robots and human swimmers, rendering traditional land-based electromagnetic communication technologies ineffective \cite{Birk_2022_CRR,Chen2016TCST}. This limitation presents substantial challenges for human-robot communication in aquatic environments.

Current underwater Human-Robot Interaction (HRI) primarily relies on acoustic sensing \cite{DeMarco2013SMC} and visual systems \cite{Chavez_2021_RAM}. However, both methods face significant limitations in complex aquatic environments. Acoustic sensing is susceptible to interference and may disrupt marine ecosystems, while visual systems are hindered by restricted fields of view and sensitivity to lighting conditions and water turbidity. Additionally, the spatial and energy requirements of these sensors impose further constraints on small underwater robots.

In contrast, the lateral line system, a crucial sensory mechanism in fish, plays a vital role in their navigation and behavior in complex underwater environments \cite{Partridge_1980_JCP,Zhai_JBE_2021,Liu_ABB_2016}. Inspired by this biological system, ALLs can detect pressure distributions and water vibrations, providing robust environmental data that remains unaffected by hydrodynamic, optical, or auditory variations. This capability has led to significant advancements in robot interaction using ALLs \cite{Zheng_2017_BIB,Yen_2020_BIB,Zheng_2020_BIB}. The energy efficiency and compact design of ALLs make them particularly suitable for deployment on small underwater robots, highlighting their substantial potential for enhancing HRI in aquatic environments.
\begin{figure}[htp!]
    \centering
    \includegraphics[width=\linewidth]{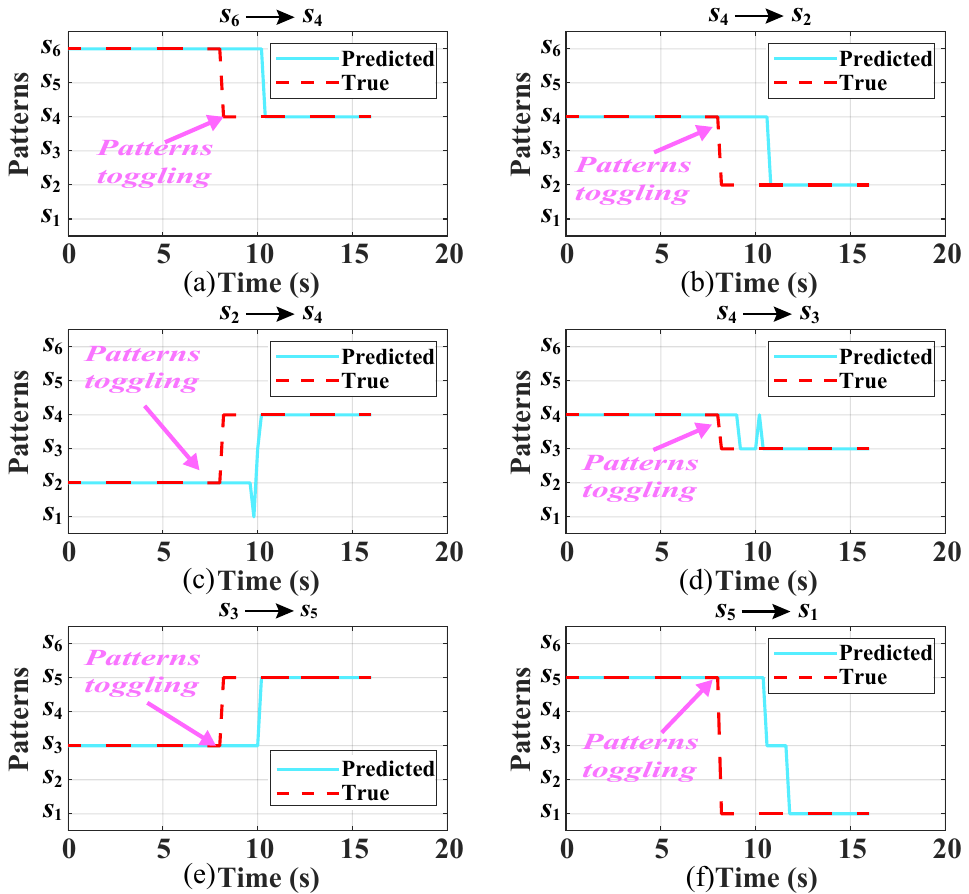}
    \caption{The experimental results of recognizing changing kick patterns.}
    \label{fig:patterns}
\end{figure}
\subsection{Performance in Recognizing Changing Kick Patterns}

Accurate identification of changing patterns is a critical aspect when applying our algorithm to real-world HRI scenarios. Experiments were designed and conducted to investigate its performance. Fig.~\ref{fig:patterns} presents the results of pattern transitions experiments, where Figs.~\ref{fig:patterns}(a)-(f) illustrate the transitions $[s_6 \to s_4, s_4 \to s_2, s_2 \to s_4, s_4 \to s_3, s_3 \to s_5, s_5 \to s_1]$. 

The results indicate that the proposed method is able to accurately identify changing patterns, while requiring a transient time of approximately 2-3 seconds. We conjecture this delay comes from the time required for the leg model to generate a new flow field after receiving the change command, as well as delayed data updates within the sliding window.

\section{Conclusion}
This paper presented a bioinspired sensing framework designed to investigate the undulatory flow field generated by leg kicks during swimming using the ALL system. By dynamically fusing time-domain and time-frequency features, the proposed method demonstrated notable advancements in identifying leg kick patterns and estimating relative positions between the leg model and ALL system. This study represented a significant step forward in bioinspired ALL sensing, providing a systematic research approach for ALL-based underwater human-robot interaction.

In future work, we will extend the proposed sensing algorithm to real-world three-dimensional environments to capture the undulatory flow field generated by real human swimming. In addition, we plan to explore leader-follower formation between underwater robots equipped with ALL systems and human swimmers, aiming to assist human swimmers in underwater tasks. 

\bibliographystyle{IEEEtran}
\bibliography{main}

\end{document}